\documentclass{article}
\usepackage{ijcai22}

\usepackage{times}

\usepackage{soul}
\usepackage{url}
\usepackage[hidelinks]{hyperref}
\usepackage[utf8]{inputenc}
\usepackage[small]{caption}
\usepackage{graphicx}
\usepackage{amsmath}
\usepackage{booktabs}
\usepackage{algorithm}
\usepackage{algorithmic}

\newtheorem{theorem}{Theorem}

\input{Definitions.sty}
\usepackage{natbib}
\newcommand{\gw}{\mathsf{GW}}
\newcommand{\fgw}{\mathsf{FGW}}
\usepackage{color}
\usepackage{cleveref}
\usepackage{subcaption}
\usepackage{multirow}
\usepackage{multibib}
\usepackage{multicol}

\newcommand{\todo}[1]{{\color{red} TODO: #1}}

\newcommand{\xun}[1]{{\todo {\color{red} Xun says: #1 }}}

\title{Gromov-Wasserstein Discrepancy with Local Differential Privacy \\for Distributed Structural Graphs}

\author{
Hongwei Jin$^1$\footnote{Contact Author}\and
Xun Chen$^2$\\
\affiliations
$^1$University of Illinois at Chicago\\
$^2$Samsung Research America, inc.\\
\emails
hjin25@uic.edu,
xun.chen@samsung.com
}

\begin{document}

\maketitle

\begin{abstract}
  Learning the similarity between structured data, especially the graphs, is one of the essential problems.
  Besides the approach like graph kernels,
  Gromov-Wasserstein (GW) distance recently draws a big attention due to its flexibility to capture both topological and feature characteristics,
  as well as handling the permutation invariance.
  However, structured data are widely distributed for different data mining and machine learning applications.
  With the privacy concerns,
  accessing the decentralized data is limited to either individual clients or different silos.
  To tackle these issues, we propose a privacy-preserving framework to analyze the GW discrepancy
  of node embedding learned locally from graph neural networks in a federated flavor,
  and then explicitly place local differential privacy (LDP) based on Multi-bit Encoder to protect sensitive information.
  Our experiments show that,
  with strong privacy protection guaranteed by $\varepsilon$-LDP algorithm,
  the proposed framework not only preserves privacy in graph learning,
  but also presents a noised structural metric under GW distance,
  resulting in comparable and even better performance in classification and clustering tasks.
  Moreover, we reason the rationale behind the LDP-based GW distance analytically and empirically.

  %

\end{abstract}

\section{Introduction}

 A lot of applications in machine learning are extensively using either pairwise matrices or kernels to perform the task of prediction or regression.
 Examples like the RBF kernel and covariance matrices are commonly used in support vector machine classification \cite{chang2010training} in the Euclidean space.
 However, for the non-Euclidean domains,
 presented in graphs,
 such as social network \citep{fan2019graph},
 recommendation systems \citep{wu2020graph},
 fraud detection \citep{li2020flowscope}
 and biochemical data \citep{coley2019graph}
 and topology-aware IoT applications \citep{abusnaina2019adversarial},
 involve the study the of relations among the featured objects.
 To measure the graph similarities, a variety of graph kernels \citep{borgwardt2005shortest, vishwanathan2010graph}
 are proposed but mainly focus on the topological information.
 Graph neural networks (GNNs, \citet{kipf2016semi, hamilton2017inductive}) are one of the state-of-the-art models to learn node embedding.
 However, it requires additional readout layers to match the dimension for graph sizes,
 and it lacks the trackable way to capture the isometric issue on graphs.
 \begin{figure}[t]
       \centering
       \includegraphics[width=0.5\textwidth]{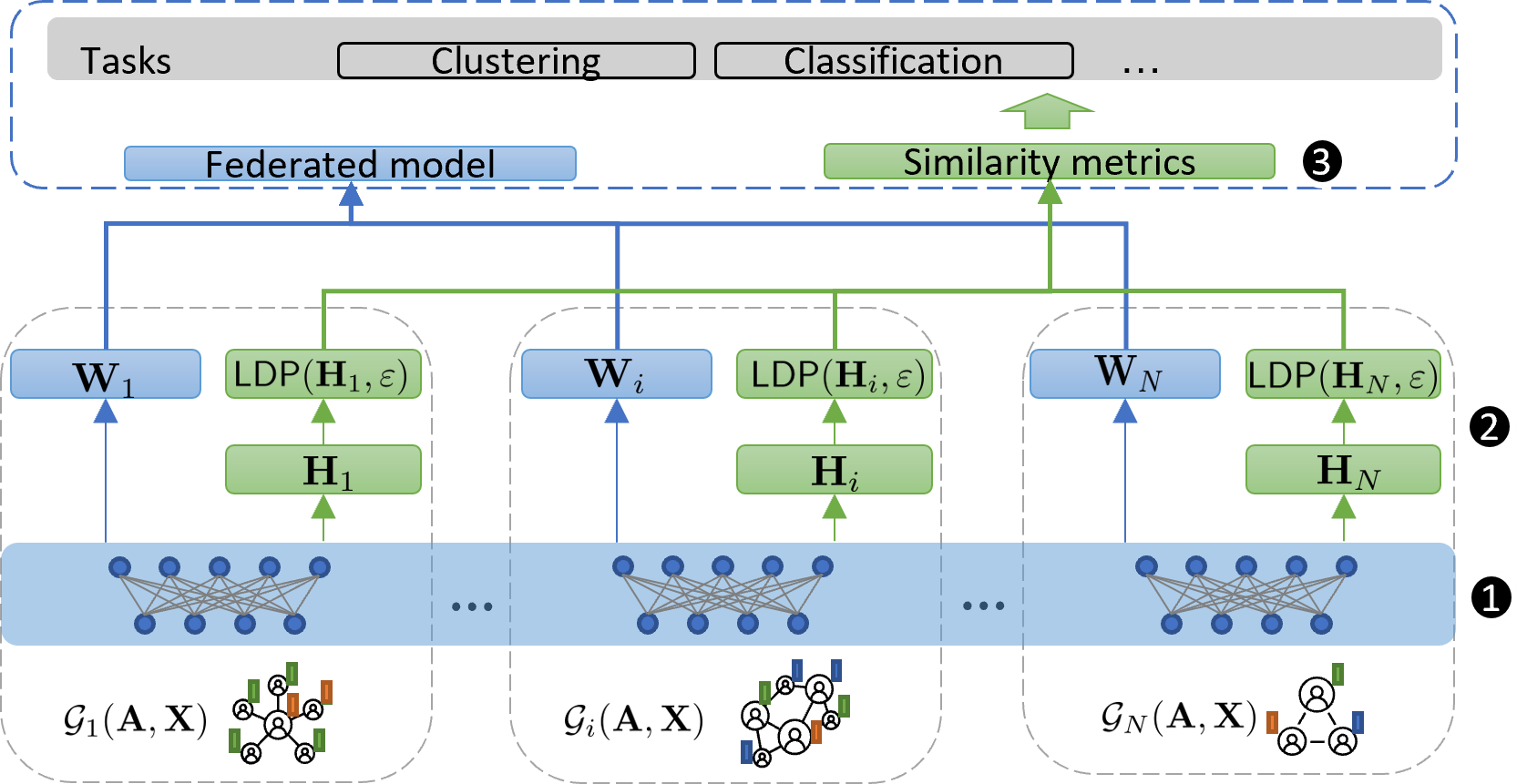}
       \caption{Framework}
       \label{fig:framework}
 \end{figure}
 Recently, Gromov-Wasserstein (GW) distance \citep{memoli2011gromov,peyre2016gromov}
 provides a geometry-based metric to measure the optimal transportation from one structural object to another.
 GW operates on the metric-measure spaces, \ie, a Borel probability measure on the set with its compact metric space.
 In other words, it operates on distances between pairs of points calculated within each domain
 and measures how the distance is compared to those in the other domain.
 Specifically, all-pair shortest path (APSP) is one of the common geodesic measures applied.
 Thus, it requires calculating the transportation between two domains by their intra-distance,
 turning the distance problem from a linear optimization to a quadratic one.
 And it extends both topological and features information (known as FGW \citep{titouan2019optimal}) so that the similarity problem cooperates with the node features nicely.
 Noting that GW is highly related to graph matching problem, encoding structural information to compare graphs, and has long history been successfully adopted in
 image recognition \citep{peyre2016gromov},
 bio-domain adoption \citep{demetci2020gromov},
 and point-cloud data alignment \citep{memoli2006computing}.

 One of the key advantages of applying GW distance is the isometric properties between structures.
 All previous works have assumed that all structured data are always available and accessible, which is often not true in reality.
 Structured data in many applications either contain sensitive and private information (such as payment transactions and social connections) or distribute across different clients commonly \citep{liao2021information, tang2020privacy}.
 On one hand, data owners are limited to sharing the original data due to the privacy regulations that are enforced globally,
 such as GDPR \citet{regulation2018general} and
 CCPA \citep{bukaty2019california}.
 On the other hand, in many applications,
 one has to deal with very large datasets that cannot be processed at once.
 Thus, a nested partitions of the input which are parameterized by a scale parameter is arising from the metric structure of the input data.

 \subsection{Problem Formulation}

  In this paper, we work on the problem of GW distance between distributed graphs while having privacy constraints.
  Two key assumptions are placed in the underlying such scenarios:
  1) each graph holds its topological information, nodes with feature and labels locally;
  2) server cannot access raw data globally while having to distinguish from graph to graph
  (anomaly detection, clustering, recommendation, etc).
  Accordingly, two main issues need to be addressed:
  first, to train the global model across different graphs while preserving the privacy of local data;
  second, to compare graphs even in the case of permutation invariant and size variant.
  In order to overcome the two issues, we take federated learning schema as our primary framework,
  along with a geometry-based metric named Gromov-Wasserstein discrepancy \citep{peyre2016gromov}
  on the server-side to learn similarity / dissimilarity across graphs.
  Gromov-Wasserstein distance \citep{memoli2011gromov} provides a metric to measure the optimal transportation from one structural object to another.


  The proposed framework is illustrated in Figure \ref{fig:framework}.
  To summarize:
  1) the synchronized vertical federated learning first updates local parameters $\Wvec$ from the private structural data $\Gcal(\Avec, \Xvec)$
  to learn the local graph embedding $\Hvec$;
  2) then, instead of pulling the raw graph data from the client,
  the proposed framework takes the local graph embedding learned from the FL model,
  and local differential privacy (LDP) is applied explicitly,
  where we extend the multi-bit encoder from vector space to matrix scenario,
  protecting the sensitive information;
  3) after getting the encoded graph embeddings,
  the framework calculates similarity metrics (GW distance) on the server and serves the downstream tasks,
  such as clustering, classification, and matching problems.



  Overall, we highlight our core contributions as
  \begin{itemize}
        \item Beyond the conventional federated learning, we design the guaranteed $\varepsilon$-LDP on graph embeddings to explicitly protect the graph information.
              It is worth mentioning that LDP on topology is not applicable because of the vulnerability of GNN,
              and perturbation on either the node or edge leads to a non-smoothing function.
        \item Besides, we adopt the Gromov-Wasserstein discrepancy to be the learnable structural metric on the server,
              cooperating the LDP encoded graph embeddings from the client-side.
        \item Interestingly, we observe that the LDP on graph embeddings not only protects privacy
              but also helps to escape saddle points in GW optimization.
              We interpret that this is largely due to the non-convexity of GW discrepancy.
        \item Having the GW discrepancy with noised input, we empirically evaluate the performance of graph classification and clustering problem.
  \end{itemize}
  To the best of our knowledge,
  this is the first work to learn the structural metric with local differential privacy under the distributed setting.
  And the framework is flexible to learn the node embedding 
  and can be combined with a variety of GNNs for a broad range of different tasks, including graph classification, clustering, etc.

 \subsection{Notation}
  We take capital and lower letter in bold to represent matrix and vector, respectively.
  A graph $\Gcal(\Avec, \Xvec)$ has adjacency matrix $\Avec$ as structural information
  and node feature $\Xvec$ as feature information.
  Besides, $\Vcal, \Ecal$ are the node sets and edge sets from the graph, respectively.
  $\Hvec$ is the retrieved node representation,
  each row of the matrix represents a single node on the graph.
  An upper script $(i)$ means the $i$-th graph, while the lower script $i$ means $i$-th row of the matrix.



\section{Related Work}
 The ultimate goal of our proposed method is to protect the privacy of structured data
 while building the GW distance for similarities between graphs.
 Naturally,
 Federated Learning (FL) \citep{mcmahan2017communication, smith2017federated}
 is an emerging collaborative learning framework that can train multiple clients locally
 while maintaining a global model on the server.
 Distinguished from the shared parameter server \cite{li2014scaling},
 one of the significant fruits of federated learning is privacy protection,
 where each client holds its raw data locally, only shared parameter/gradient will be exchanged with the server.
 Specifically, in the vertical FL setting, a model is shared among all the clients,
 and the client trains the model locally with its labels,
 while the server learns the global models by aggregating the local client models. 

 Recently, a set of different methods are proposed to build a GNN model with a federated setting.
 %
 %
 With the task of graph classification,
 \citet{he2021fedgraphnn} design a federated learning system and benchmark for graph neural networks.
 The framework is built based on FedML \citep{he2020fedml} research library,
 enabling to run a variety of GNN models effectively with non-IID graph dataset,
 which is a similar approach in the image domain.
 Interestingly, FL-AGCNS \citep{wang2021fl} propose to take advantage of federated learning
 to accelerate in the task of neural architecture search of graph convolutional network.
 Specifically, they take federated SuperNet optimization to select random GCN architecture
 and take federated evolutionary optimization to perform evaluation suing GCN SuperNet.


 However, all the work above trains a shared model with the task of supervised/semi-supervised learning,
 but do not address the issue of unlabelled graphs on clients.
 What's more, none of the work above discusses the graph differential privacy while learning the model.
 Unlike our proposed framework, a recent work of locally private graph neural network \citep{sajadmanesh2020locally} build the feature and label differential private on the clients at once while leaving the whole training process on the server-side instead.

 Another approach to protecting privacy is differential privacy.
 More specifically, with distributed graphs,
 we mainly focus on the local differential privacy of each graph.
 Given a graph $\Gcal(\Avec, \Xvec)$, the purpose of applying LDP is to protect the private information from topology $\Avec$ and node feature $\Xvec$. A naive approach would be directly adding LDP noise to $\Avec$ and $\Xvec$. However, we argue that such an approach has its limitations.
 For the topological LDP, assume we add random perturbation to the topology of the graph, \ie, add or delete edges on a graph.
 However, works like Nettack proposed in \citet{zugner2018adversarial},
 found that even a small perturbation could lead to the misclassification of the nodes in the graph.
 Moreover, random perturbation could result in the same graph but is permutation invariant to the original graph,
 \ie, $\Avec' = \Pi(\Avec)$, where $\Avec'$ is the perturbed graph \citep{niu2020permutation}.
 For the feature LDP, adding LDP on feature is a valid approach in the work \citet{sajadmanesh2020locally}, but it lacks protection for the structural information. Moreover, the study in graph diffusion Wasserstein distance \citep{barbe2020graph} suggests that a noisy feature will let the fused GW depend mostly on the structure.
 Therefore, adding small random perturbation to either the topology or the feature is not a guaranteed approach to protect privacy in the federated learning setting.

 \subsection{Gromov-Wasserstein Discrepancy}
  Graph neural networks rely on training on structured data for various graph-related tasks.
  However, a common limitation is its difficulty in explaining the isomorphism in graph data.
  Identification of similarities between graphs is an essential problem in graph learning areas.
  In these areas, the graph isomorphism problem is known as the exact graph matching \citep{xu2019gromov},
  which is not solvable in polynomial time nor to be NP-complete.
  Two graphs are considered {\it isomorphic} if there is a mapping between the nodes of the graphs that preserves their adjacencies.
  The graph isomorphism testing is used in a wide range of applications,
  such as the identification of chemical compound \citep{demetci2020gromov},
  the generation of molecular graphs in chemical dataset \citep{titouan2019optimal},
  and the electronic design automation (EDA) with placement and routing operations \citep{chan2000multilevel}.
  Graph Isomorphism Network (GIN) \citep{xu2018powerful} is recently proposed
  to implement Weisfeiler-Lehman (WL) graph isomorphism test \citep{shervashidze2011weisfeiler}.
  However, such kind of approach only deals with graphs of the same size,
  and hard to distinguish the difference between graphs with arbitrary sizes.

  Recently, the optimal transport (OT) associated with their Gromov-Wasserstein (GW) discrepancy \citep{peyre2016gromov},
  which extends the Gromov-Wasserstein distance \citep{memoli2011gromov},
  has emerged as an effective transportation distance between structured data,
  alleviating the incomparability issue between different structures by aligning the \textit{intra}-relational geometries.
  GW distance is isometric, meaning the unchanged similarity under rotation, translation, and permutation.
  %
  Formally, \citet{peyre2016gromov} define the Gromov-Wasserstein discrepancy between two measured similarity matrices
  $(\Cvec, \pvec) \in \RR^{n\times n} \times \sum_n$ and $(\Dvec, \qvec) \in \RR^{m \times m} \times \sum_m$ as follows
  \begin{align}
    \gw (\Cvec, \Dvec, \pvec, \qvec)
    = \min_{\Tvec \in \Ccal_{\pvec, \qvec}} \sum_{i, j, k, l} \ell (\Cvec_{i, k}, \Dvec_{j, l}) \Tvec_{i, j} \Tvec_{k, l},
  \end{align}
  where $\Cvec$ and $\Dvec$ are matrices representing either similarities or distances between nodes within the graph.
  $\ell(\cdot)$ is the loss function either in Euclidean norm or KL-divergence \citep{peyre2016gromov}.
  $\pvec \in \RR^+_n: \sum_i \pvec_i = 1$ is the simplex of histograms with $n$ bins,
  and $\Tvec$ is the coupling between the two spaces on which the similarity matrices are defined.
  Specifically,
  \begin{align*}
    \Ccal_{\pvec, \qvec} = \cbr{\Tvec \in \RR_+^{n \times m}, \Tvec \one = \pvec, \Tvec^\top \one  = \qvec},
  \end{align*}
  here, we denote the domain $\Ecal := \cbr{\Tvec \in \RR^{n \times m}, \Tvec \one = \pvec, \Tvec^\top \one  = \qvec}$
  and $\Ncal := \RR_+^{n \times m}$.
  Furthermore, we can rewrite the problem in Koopmans-Beckmann form \citep{koopmans1957assignment}:
  \begin{align}
    \label{eq:gw_form2}
    \mathsf{GW} (\Cvec, \Dvec, \pvec, \qvec) = \frac{\norm{\Cvec}_F^2}{n^2} + \frac{\norm{\Dvec}_F^2}{m^2}
    - 2 \max_{\Tvec \in \Ccal_{\pvec, \qvec}} \tr (\Cvec \Tvec \Dvec \Tvec^\top).
  \end{align}
  Note that the $\gw$ distance is non-convex and highly related to the quadratic assignment problem (QAP), which is NP-hard.
  Therefore, given the similarity matrices $\Cvec$ and $\Dvec$,
  we optimize with the quadratic trace form in \eqref{eq:gw_form2} over the domain of $\Ecal \cap \Ncal$.
  The GW discrepancy problem can be solved iteratively by conditional gradient method \citep{peyre2016gromov}
  and the proximal point algorithm \citep{xu2019scalable}.


  In the next section, we will address the details of our proposed $\varepsilon$-LDP on graph embeddings and cooperate them under GW metrics.

\section{Framework}
 To highlight our framework, it encodes node representation from the client while building the structural metric on the server-side.
 Regarding build the node representation, we train the model in the task of node classification,
 by using the node labels locally.
 A conventional federated learning framework is well suited in this setting (detailed description in Appendix~\ref{sec:fl}).
 To the simplification, we take FedAvg as the aggregation algorithm without further notice.
 Besides, we apply a multi-bit encoder to the node representation as LDP before transmit the graph information to the server.
 With the protected node representation, server calculates the geometric metric GW accordingly.



 \subsection{$\varepsilon$-LDP with Multi-bit Encoder on Graph Embedding}
  Inspired by the locally differential privacy on features, we adopt the multi-bit encoder from \citet{sajadmanesh2020locally},
  while their approach send the encoded feature and label directly to server and train a single model on the server-side.
  Differentiated from feature encoding, we take the encoder on the node representation, leveraging the privacy on both topology and feature learned from GNN.
  Algorithm \ref{alg:multi-bit-encoder} describes the LDP on node embedding in more details.
  We basically extend the one-bit encoder on a vector space to the multi-bit encoder in 2-D dimension while preserving the guaranteed privacy.
  Intuitively, the encoder first uniformly sample the $m$ out of the dimension of $n \times h$ without replacement.
  Then for each row-wise dimension, we take the Bernoulli distribution to decide the mapped encoding after the mask.
  For each hidden feature dimension, the probability encoding is determined by the number of nodes in the graph ($n$) and the sampled mask ($m$).
  This is just one-pass encoding with the complexity of $\Ocal(m n)$.

  \begin{algorithm}
    \caption{$\mathsf{LDP}_{G}(\cdot, \varepsilon)$: Multi-Bit Encoder}
    \label{alg:multi-bit-encoder}
    \textbf{Input}: Node embedding $\Hvec \in \sbr{\alpha, \beta}^{n \times h}$, $\varepsilon$ \\
    \textbf{Output}: Encoded embedding $\Hvec^*$
    \begin{algorithmic}[1]
      \STATE Let $\Scal$ be a set of $m$ values sampled uniformly from $\sbr{n} \times \sbr{h}$
      \FOR{$i$ in $n$}
      \FOR{$j$ in $h$}
      \STATE $s_{ij} = 1$ if $\Scal_{ij}=1$ else $s_{ij} = 0$
      \STATE $t_{ij} \sim \text{Bernoulli} \rbr{\frac{1}{e^{\varepsilon/(mn)} + 1} + \frac{\Hvec_{ij} - \alpha}{\beta - \alpha} \cdot \frac{e^{\varepsilon/(mn)} - 1}{e^{\varepsilon/(mn)} + 1}}$ \\
      \STATE $\Hvec^*_{ij} = s_{ij} (2 t_{ij} - 1)$
      \ENDFOR
      \ENDFOR
    \end{algorithmic}
  \end{algorithm}

  \begin{theorem}
    \label{thm:dp}
    The multi-bit encoder presented in Algorithm \ref{alg:multi-bit-encoder} satisfies $\varepsilon$-local differential privacy for each graph representation.
  \end{theorem}
  Theorem \ref{thm:dp} provides the theoretically guaranteed privacy protection on hidden representation.
  A detailed proof is provided in the appendix \ref{proof:thm_dp}.

 \subsection{$\gw$ Distance with LDP on Graph Embeddings}
  Besides the work of building the global model,
  we also explicitly enforce the server to construct a pairwise similarity matrix $\Cvec \in \RR^{n \times n}$.
  More specifically, $\Cvec_{\Hvec^{*(i)}} \in \RR^{n^{(i)} \times n^{(i)}}$ represents intra-distance of the encoded node embedding from $i$-th client,
  namely $\Hvec^{*(i)} \in \RR^{n^{(i)} \times h}$, where $n^{(i)}$ is the number of node in $i$-th graph and $h$ is the hidden dimension.
  Therefore, we can explicitly formulate the distance matrix between node $u$ and $v$ in $\Cvec$ as
  \begin{align}
    \label{eq:gw_h}
    \Cvec_{\Hvec^*}(u, v) = d(\Hvec_u^*, \Hvec_v^*), \  \forall u, v \in \Vcal,
  \end{align}
  where $d(\cdot, \cdot)$ is the distance function applied.
  Due to its Euclidean space over the reals of $\Hvec^*$,
  we can simply take the Euclidean distance between each node embedding to build intra-distance.
  Of course, it can be easily extended to other distance metrics, such as cosine similarity and correlation.
  Noting that, after taking the distance introduced from hidden representation,
  the $\Cvec_{\Hvec}$ is a metric space induced from the original graph (without $\varepsilon$-LDP encoding).
  Even though the unbalanced probability measure over nodes are also applicable \citep{titouan2019optimal},
  we will take $\pvec$ as a uniform distribution among the nodes, \ie, $\pvec_{u} = \frac{1}{\abs{\Vcal}}, \forall u \in \Vcal$.
  Also, as $\Hvec$ is the graph representation after federated learning,
  the embedding itself encodes both structural ($\Avec$) and feature $(\Xvec)$ information,
  which is different from the conventional calculating $\Cvec$ where only geodesic distance is applied commonly.

  Additionally, the framework requires the optimization of Gromov-Wasserstein discrepancy on the server,
  and the $\varepsilon$-LDP encoded graph embedding can be retrieved only once after a stable global model from FL learning.
  However, considering the unexpected drop-off of clients, the graph embedding can be retrieved periodically if necessary.
  Therefore, compared with the vanilla FL setting,
  sending the encoded graph embedding $\Hvec^*$ will not introduce overhead in communication.

\section{Experiments}

 \paragraph{Dataset}
 We will perform two tasks, one with graph classification under federated learning via GW,
 where we take the benchmark datasets from TUDataset \citep{morris2020tudataset}.
 Another task is to take the graph clustering with subgraph generated from the single big graph,
 where we take the citation benchmark dataset from Planetoid \citep{yang2016revisiting}.
 Detailed statistics of the dataset and settings of hyperparameters are included in the appendix \ref{sec:exp_setup}.

 \subparagraph{Graph classification}

 \begin{table}
       \centering
       \resizebox{0.5\textwidth}{!}{
             \begin{tabular}{|l|c|c|c|c|}
                   \hline
                   Model & Metric Source                 & PROTEINS         & MUTAG            & IBMD-B           \\
                   \hline
                   SPK   & $\Avec$                       & n/a              & 82.95 $\pm$ 8.19 & 55.80 $\pm$ 2.93 \\
                   \hline
                   GCN   & $\Avec \Xvec$                 & 69.03 $\pm$ 3.59 & 80.05 $\pm$ 3.21 & n/a              \\
                   \hline
                   SVM   & $\gw_\Avec/\fgw_{\Avec\Xvec}$ & 74.55 $\pm$ 2.74 & 83.26 $\pm$ 10.3 & 63.8 $\pm$ 3.49  \\
                   \hline\hline
                   SVM   & $\gw_{\Hvec (GCN)}$           & 72.31 $\pm$ 4.61 & 80.3 $\pm$ 9.75  & 53.69 $\pm$ 4.31 \\
                   \hline
                   SVM   & $\gw_{\Hvec^* (GCN)}$         & 71.19 $\pm$ 3.92 & 79.57 $\pm$ 6.73 & 54.01 $\pm$ 4.90 \\
                   \hline
                   SVM   & $\gw_{\Hvec (GIN)}$           & 75.53 $\pm$ 2.01 & 93.08 $\pm$ 3.35 & 74.4 $\pm$ 1.11  \\
                   \hline
                   SVM   & $\gw_{\Hvec^* (GIN)}$         & 76.89 $\pm$ 3.90 & 91.17 $\pm$ 4.21 & 73.6 $\pm$ 3.23  \\
                   \hline
             \end{tabular}
       }
       \caption{Accuracy on graph classification}
       \label{tab:glaph_classification}
 \end{table}

 Regarding the task of graph classification under a centralized setting,
 we adopt an SVM using the indefinite kernel matrix $e^{-\gamma \fgw}$ as a distance between graphs \citep{titouan2019optimal},
 where the training, validation, and testing sets are split in the ratio of $7:2:1$.
 For all methods using SVM, we cross validate the parameter $C\in \cbr{10^{-7}, 10^{-6}, \cdots, 10^{7}}$ and $\gamma \in {2^{-10}, 2^{-9}, \cdots, 2^{10}}$.
 For the decentralized setting, we split the graphs into 10 clients from the Dirichlet distribution \citep{wang2020graphfl},
 and each client has 20\% and 10\% for validation and testing, respectively.
 We report the classification accuracy in Table~\ref{tab:glaph_classification}.
 Both PROTEINS and MUTAG has node features, where we adopt the $\fgw$ as the base metric.
 While IMDB-B has no features on nodes, we only take the GW (topology only) as the base metric.
 SPK~\cite{borgwardt2005shortest} is the shortest path kernel method on raw data.
 GCN indicates a two-layer graph convolutional layer with hidden dimension of 16, followed by an average pooling method.
 GIN represent a five graph isomorphic layers with hidden dimension of 64, followed by a two linear perception layers.

 Comparing the results from raw data (metric based on $\Avec, \Xvec$),
 we can see the kernel based SVM on GW discrepancy performs better than the base GCN and SPK model.
 For the FL setting, we defined the shared model with GCN and GIN.
 $\Hvec$ is the graph embedding without privacy protection, and $\Hvec^*$ is $\varepsilon$-LDP graph embedding.
 Without the further notice, we take the optimal $m^* = \max(1, \min(d, \floor{\frac{\varepsilon}{2.18}}))$ from \citet{sajadmanesh2020locally},
 and $d$ is the number of classes from nodes.
 We exact the node embedding based after the softmax layer, leaving $\alpha=0, \beta=1$.
 The default value is $\varepsilon = \smallfrac{1}{\abs{\Vcal}}$ which varies from each graph.
 Intuitively, this basically means the cumulative privacy budget will be normalized on the number of nodes $n$.

 To see the effectiveness of FL, GCN model under FL perform a bit worse than SVM on raw data.
 However, FL under GIN (a strong model) improves almost 10\% when compared with vanilla SVM with raw data.
 What's more, to check how LDP works (using the $\Hvec^*$ graph embedding) in the classification tasks,
 we can find that the $\varepsilon$-LDP didn't hurt the overall performance.

 \vspace{-1em}
 \subparagraph{Graph clustering}
 Our framework fits unsupervised learning with federated learning.
 And it is quite common to see that labels for each node are trivial, but not for the subgraphs (communities).
 Regarding the data with a single graph,
 we split the big graph into small subgraphs by taking multi-hop neighbors for simplification.
 Moreover, for such kind of dataset, we might not have the ground-truth label of the split subgraph.
 Therefore, we take the label of the central node as its pseudo-label.
 To note, we will see the overlap of nodes in different clients, which is a typical scenario in various applications, such Internet-of-Things (IoT) devices, community connections, and recommendation systems.
 A shared 2-layer GCN model with node classification task is deployed in the federated setting, where each client builds the node embedding.
 We retrieve the node embedding after a fixed number of 200 rounds of updating the shared model.

\begin{figure*}[t]
       \centering
       \includegraphics[width=0.9\textwidth]{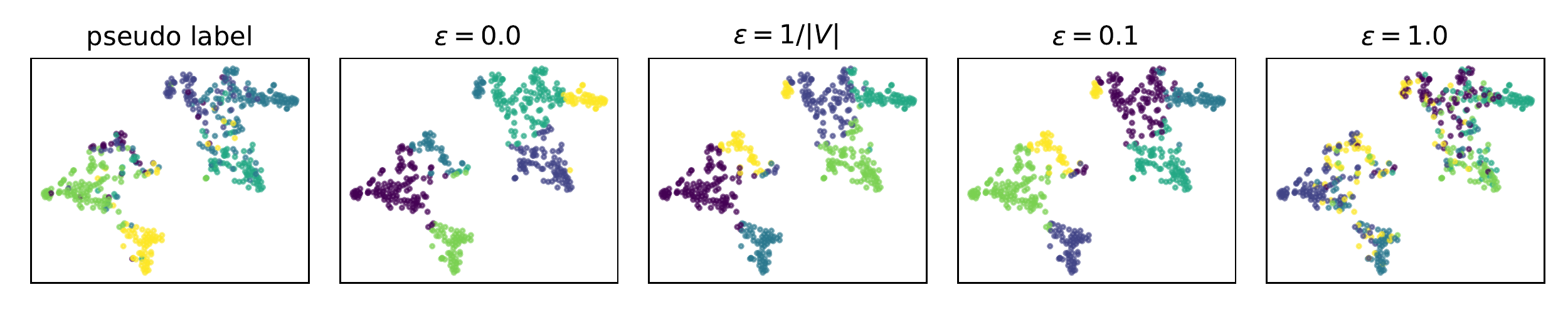}
       \caption{Clustering on subgraphs based on $\gw$. Pseudo label comes from the central node.}
       \label{fig:kmeans_hop1}
 \end{figure*}

 Figure~\ref{fig:kmeans_hop1} shows the t-SNE plot of KMeans clustering of 500 1-hop subgraphs from citeseer dataset with different $\varepsilon$.
 KMeans under GW distance is different from Euclidean distance, the centroid of each cluster is represented by a metric matrix.
 To simplify, we fix the dimension of the centroid by the average number of nodes within the cluster.
 Noting that $\varepsilon=0$ means no LDP introduced to the graph embedding and we assume the number of clusters is equal to the number of labels from nodes.
 Intuitively,
 data in the citation network has the label for each paper,
 and the induced subgraph is the citations referred for the central node.
 To see the impact of GW distance in clustering,
 comparing the first and second plots in Figure~\ref{fig:kmeans_hop1} shows that even without the LDP encoder,
 the KMeans clustering with GW distance performance better than the naive labeling from central node.
 Thanks to the property of GW distance, it provides a better similarity metrics,
 especially having a pre-learned graph embeddings.
 To see the impact of LDP in clustering, compared with the raw graph embedding,
 the encoded $\Hvec^*$ with small $\varepsilon$ won't hurt too much in the clusters.
 However, a larger privacy strength ($\varepsilon=1$) introduces more deviates of unlabelled subgraphs.
 This is consistent with the observation in classification problem as well.
 Additional results with 2-hop subgraphs and other datasets are provided in Appendix~\ref{sec:exp_setup}.
 In the next subsection, we will have a close look at privacy strength $\varepsilon$ in the GW distance calculation.


 \subsection{Privacy analysis for LDP on embedding}
  \begin{figure}[h]
        \centering
        \includegraphics[width=0.21\textwidth]{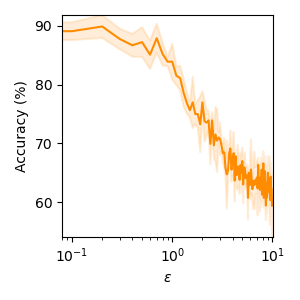}
        \includegraphics[width=0.21\textwidth]{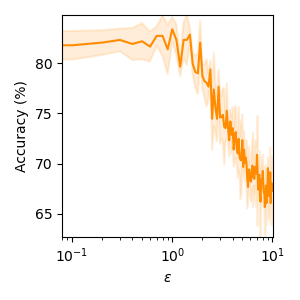}
        \caption{Cross validation on $\varepsilon$ with KNN model (left: $k=5$, right: $k=20$).
        }
        \label{fig:ldp_clf}
  \end{figure}

  \begin{figure}[h]
        \centering
        \includegraphics[width=0.36\textwidth]{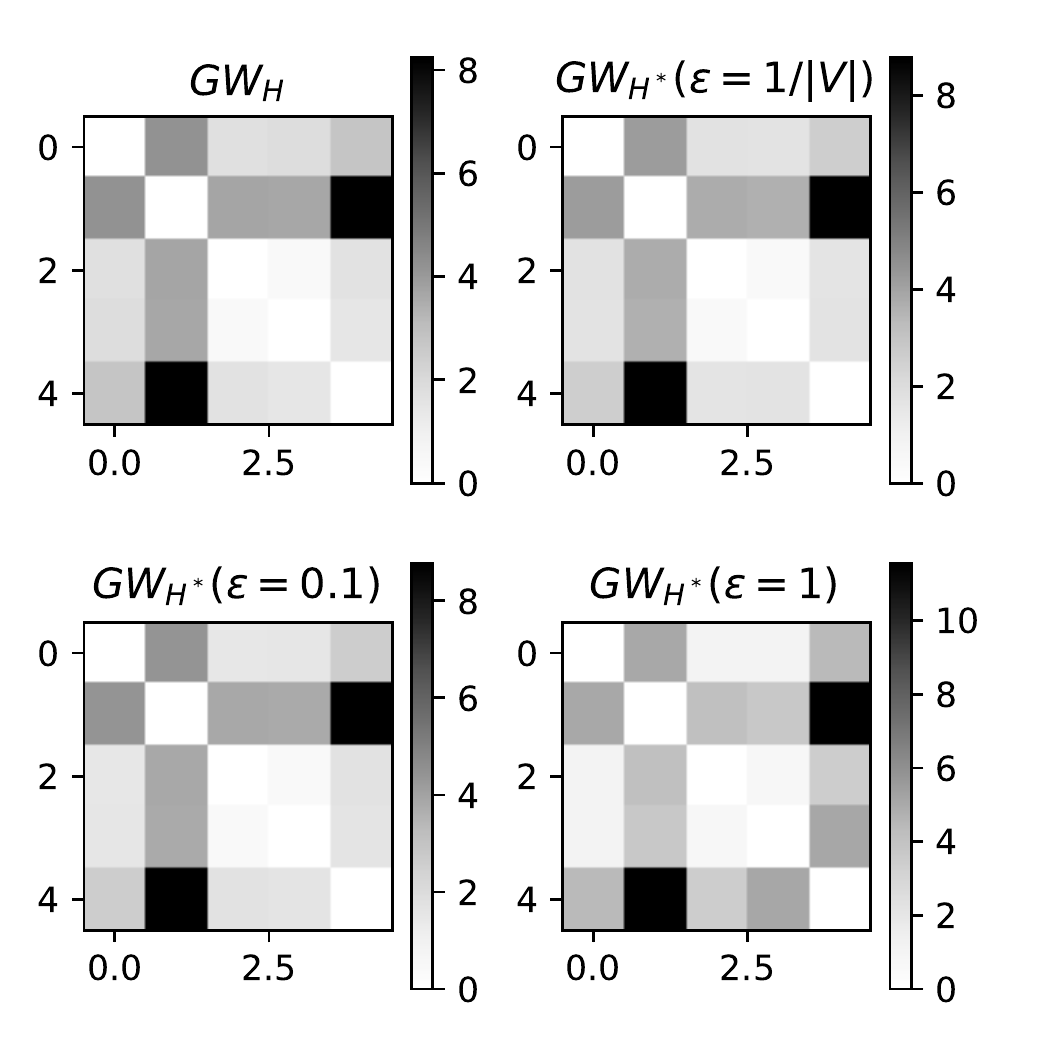}
        \caption{A demonstration of LDP on node embedding for GW metric.
        }
        \label{fig:ldp_gw}
  \end{figure}
  Knowing that the $\varepsilon$ is a key factor in privacy protection and downstream tasks,
  we invest the impact of privacy budget $\varepsilon$ in LDP under classification.
  Following the Figure~\ref{fig:ldp_clf}, in order to eliminate the influence of hyperparameter in GNN models,
  we report the accuracy using KNN based on $\gw_{\Hvec^*}$ by cross-validating on $\varepsilon$ on MUTAG dataset.
  As we can see on both settings of $k=5, 20$, the model is relative stable with just a small perturbation,
  \ie, $\varepsilon < 1$, and with the increased privacy protection strength, a larger $\varepsilon$ will hurt the overall performance.
  More interestingly, by introducing the $\varepsilon$-LDP, it not only helps to protect the privacy,
  but also potentially helps the performance in downstream tasks with a proper $\varepsilon$.

  Furthermore, we look into details of changes in similarity metric $\gw$ distance with respect to the $\varepsilon$.
  For  the task of clustering, we extract five subgraphs (check Appendix~\ref{sec:similar_metric} for subgraph structures) from citeseer dataset,
  and shows the results of $\gw$ discrepancy when LDP is applied with different $\varepsilon$ in Figure \ref{fig:ldp_gw}.
  We consider two privacy budgets $\varepsilon=0.1, 1$ to be compared.
  $\gw_H$ represents the distance without LDP, \ie, $\varepsilon=0$, with the baseline statistics of $\gw$ discrepancy to be $2.38 \pm 2.22$.
  $\gw_{H^*}$ represents the distance under encoded LDP with three
  different privacy budget $\varepsilon = \smallfrac{1}{\abs{\Vcal}}, 0.1, 1$,
  and $2.42 \pm 2.33$, $2.43 \pm 2.35$, $3.25 \pm 3.09$
  are the distances statistics, respectively.
  It can be seen from Figure~\ref{fig:ldp_gw},
  with a proper privacy budget, we can see little difference on pairwise distance based on $\gw$.
  In other words,
  from the perspective of applications, with LDP on node embedding has little affection for the clustering / classification tasks.
  It is also true from the algorithmic view that $\gw$ distance is quantitatively stable \citep{memoli2011gromov} in the sense of the structure invariant.
  %
  Moreover, in the distributed training, graphs are completely in non-i.i.d. distribution,
  suggesting that the $\varepsilon$ can vary from client to client.
  Therefore, we have our default $\varepsilon$ set to be $1/\abs{\Vcal}$,
  which also matches our design of algorithm~\ref{alg:multi-bit-encoder}, sampling based on number of nodes.
  We also provide an insightful explanation of taking GW distance when comparing with other graph similarity metrics in Appendix~\ref{sec:similar_metric}

 \subsection{Rationale behind GW with $\varepsilon$-LDP encoder}
  We step further to reason why the LDP on $\Hvec$ won't sacrifice the $\gw$ distance.
  Due to the non-convexity of GW discrepancy, which involves a quadratic problem,
  the absolute error in distance from different pairs cannot have a guaranteed boundary to tell the closeness between them.
  Despite the challenge of finding the exact the global minima,
  an effective way is to escape saddle points in the non-convex manifold,
  which prevents the optimal solution to be stuck at the local minima.
  \begin{figure}[h]
        \centering
        \includegraphics[width=0.22\textwidth]{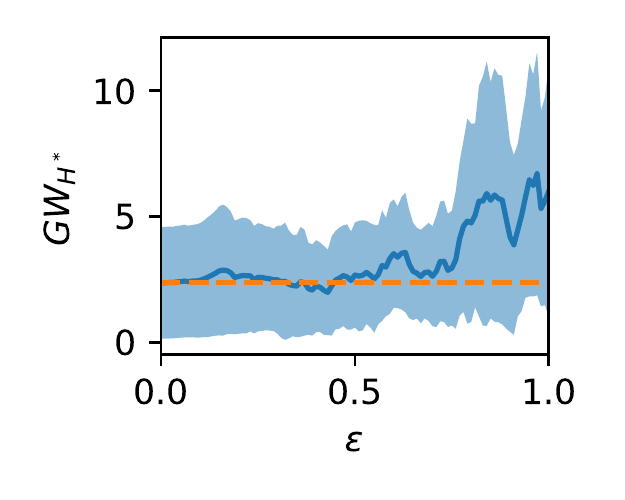}
        \includegraphics[width=0.22\textwidth]{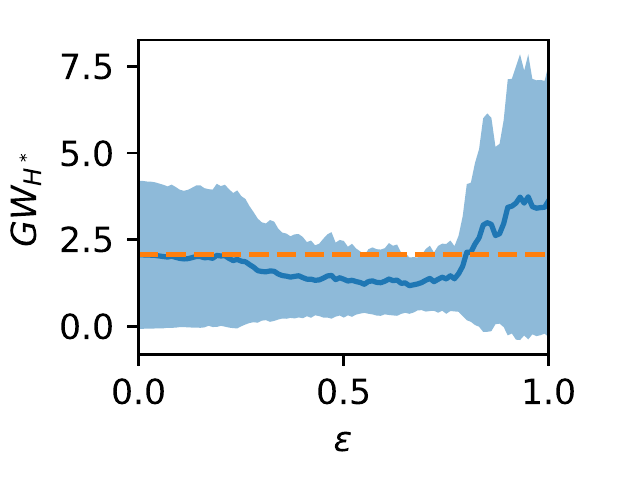}
        \caption{Averaged GW discrepancy (optimal function value) with the increase of $\varepsilon$. Left: graphs induced from 1-hop neighbors. Right: graphs induced from 2-hop neighbors.}
        \label{fig:ldp_comp}
  \end{figure}

  Among many of the works targeting the escape saddle points, perturb gradient descent (PGD) \cite{jin2017escape} is one of practical simple method to solve the problem.
  The meta-algorithm is described as
  \begin{align}
        \xvec_t \leftarrow \xvec_t + \vec{\xi}_t, \quad \quad
        \xvec_{t+1} \leftarrow \xvec_t - \eta \grad f(\xvec_t)
  \end{align}
  where $\vec{\xi}_t \sim \Bcal(\varepsilon)$ is a small perturbation on the input space within a small ball.
  Noting that, the $\Hvec^* = \Hvec + \Ocal(\varepsilon)$
  and $\Cvec_{u, v}=d(\Hvec^*_u, \Hvec^*_v) = d(\Hvec_u, \Hvec_v) + \Ocal(\varepsilon)$,
  when $d$ is a Euclidean distance function.
  Interestingly, we can take the LDP as small perturbation on the node embedding,
  which is ultimately the noisy input to the GW optimization problem.
  Lacking of exact tracking the impact of noise in optimization,
  we empirically check the average output of GW with the increase of the $\varepsilon$.
  Figure \ref{fig:ldp_comp} indicates the average GW discrepancy calculated by increasing the privacy budget $\varepsilon$ on the dataset Citeseer.
  We fix the initial point and other hyperparameters (threshold, max iter and etc.) in optimization,
  the only differences are $\varepsilon$.
  The solid blue curve indicates the average GW discrepancy (optimal function value) and the dashed orange line indicate the baseline without any perturbations.
  Although the increase of the privacy budget $\varepsilon$ eventually leads to a higher mean and variance of GW discrepancy which is expected, a slight perturbation from LDP (when $\varepsilon$ is small) helps to get a lower GW discrepancy (smaller optimal function value from blue curve), especially when the graph is more complicated (2-hop neighbors).

  The empirical result matches with our understand of non-convex optimization with perturbed method.
  However, a rigorous analysis of relation between LDP and non-convex still remains unclear,
  especially in the connection between privacy budget and the constrained perturbation domain.
  We will leave this problem for our further work.

\section{Conclusion}
 In this paper, we propose a novel framework to protect the distributed structural graphs by introducing the local differential privacy to graph embeddings
 and handle the permutation invariant issue by taking the Gromov-Wasserstein distance for further downstream tasks.
 With the guaranteed $\varepsilon$-LDP encoding on graph embedding,
 our experiment shows that the privacy budget not only helps in the distributed privacy protection learning,
 but also won't hurt the performance of the downstream tasks.
 We also invest the reasons behind it, thanks to the non-convexity of $\gw$ distance, the additional noise in LDP is not a burden in the downstream tasks.
 Considering the flexibility of the framework, an interesting extension is to see the performance on large heterogeneous graphs,
 where the subgraphs represent small communities in social networks,
 fraud actions in transactions networks, and browsing logs in recommendation systems.
   {
     \bibliographystyle{named}
     \bibliography{ijcai22}}
 \clearpage
 \appendix

\section{Definition of LDP}
 \begin{definition}[LDP \citep{dwork2014algorithmic}]
   Let $\Acal: \Dcal^n \Rightarrow \Zcal$ be a randomized algorithm
   mapping a dataset with $n$ records to some arbitrary range $\Zcal$.
   The algorithm $\Acal$ is $\varepsilon$-local differential privacy
   if it can be written as $\Acal(d^{(1)}, \cdots, d^{(n)}) = \Phi(\Acal_1(d^{(1)}, \cdots, \Acal_n(d^{(n)})))$,
   where each $\Acal_i: \Dcal \rightarrow \Ycal$ is an $\varepsilon-$local randomizer for each $i \in \sbr{n}$
   and $\Phi: \Ycal^n \rightarrow \Zcal$ is some post-processing function
   of the privatized records $\Acal_1(d^{(1)}, \cdots, \Acal_n(d^{(n)}))$.
   Note that the post-processing function does not have access to the raw data records.
 \end{definition}

\section{Proof of Theorem \ref{thm:dp}}
 \label{proof:thm_dp}
 \begin{proof}
   Let $\Mcal(\Hvec)$ to be the multi-bit encoder as described in Algorithm \ref{alg:multi-bit-encoder}.
   Let $\Hvec^*$ be the perturbed node embedding corresponding to $\Hvec$.
   We need to show that for any two embeddings $\Hvec$ and $\Hvec'$, we have
   $\ln{\frac{\PP\sbr{\Mcal(\Hvec) = \Hvec^*}}{\PP\sbr{\Mcal(\Hvec') = \Hvec^*}}} \le \varepsilon$.
   According to Algorithm \ref{alg:multi-bit-encoder}, for each row-wise vector representation, we have seen that $\Hvec_{i, :}^* \in \cbr{-1, 0, 1}$ and the case of $\Hvec_{i,:}^*=0$ occurs when $i \notin \Scal$. Therefore, we have
   \begin{align}
     \ln \frac{\PP\sbr{\Mcal(\Hvec_{i, :}) = 0}}{\PP\sbr{\Mcal(\Hvec'_{i,:}) = 0}}
     = \ln \frac{1 - (mn)/h}{1-(mn)/h} = 0
     \le \varepsilon,
   \end{align}
   for all $\varepsilon > 0$ and $i \in \sbr{n}$.
   In the case of $\Hvec_{i, :}^* \in \cbr{-1, 1}$, we see that the probability of getting $-1$ or $1$
   ranges from
   $\frac{(mn)}{h} \cdot \frac{1}{e^{\varepsilon/ (mn)} + 1}$ to
   $\frac{(mn)}{h} \cdot \frac{e^{\varepsilon/ (mn)} }{e^{\varepsilon/ (mn)} + 1}$.
   Then we have
   \begin{align*}
       & \frac{\PP\sbr{\Mcal(\Hvec_{i, :}) \in \cbr{-1, 1}}}{\PP\sbr{\Mcal(\Hvec'_{i,:}) \in \cbr{-1, 1}}}
     \le \frac{\max \PP\sbr{\Mcal(\Hvec_{i, :}) \in \cbr{-1, 1}}}{\min \PP\sbr{\Mcal(\Hvec'_{i,:}) \in \cbr{-1, 1}}}
     \\
     = & \rbr{\frac{(mn)}{h} \cdot \frac{e^{\varepsilon/ (mn)} }{e^{\varepsilon/ (mn)} + 1}} / \rbr{\frac{(mn)}{h} \cdot \frac{1}{e^{\varepsilon/ (mn)} + 1}}
     = e^{\varepsilon / (mn)}.
   \end{align*}
   Then by the composition theorem, we have
   \begin{align*}
     \frac{\PP\sbr{\Mcal(\Hvec) = \Hvec^*}}{\PP\sbr{\Mcal(\Hvec') = \Hvec^*}}
      & = \prod_{i=1}^n \prod_{j=1}^h \frac{\PP\sbr{\Mcal(\Hvec_{ij}) = \Hvec_{ij}^*}}{\PP\sbr{\Mcal(\Hvec_{ij}') = \Hvec_{ij}^*}}
     \\
      & \le \prod_{i=1}^n\prod_{\Hvec^*_{ij}\in \cbr{-1, 1}} e^{\varepsilon / (mn)} \le e^{\varepsilon},
   \end{align*}
   which concludes the proof of $\varepsilon$-d.p.
 \end{proof}

\section{More experiments}
 \label{sec:exp_setup}
 We take the experiments in two scenarios: one with single graph, but split into multiple subgraphs;
 the other takes the standard graph classification benchmark.
 Table~\ref{tab:ds1} and Table~\ref{tab:ds2} are the statistics of those datasets.
 Table~\ref{tab:subgraph} is the statistics of induced subgraphs from citation networks.

\begin{figure*}
   \centering
   \includegraphics[width=0.9\textwidth]{figures/citeseer_kmeans_graph_emb_hop_1.pdf}
   \includegraphics[width=0.9\textwidth]{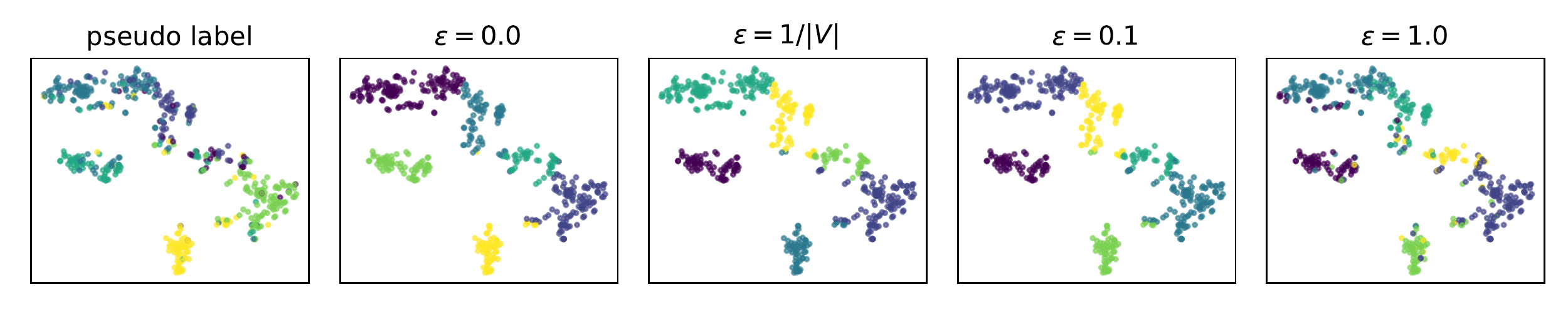}
   \caption{Clustering from 1-hop (top) and 2-hop (bottom) subgraphs of citeseer}
   \label{fig:citeseer_cluster_hop}
 \end{figure*}

\begin{figure*}
   \centering
   \subfloat[1-hop clustering]{
     \includegraphics[width=0.4\textwidth]{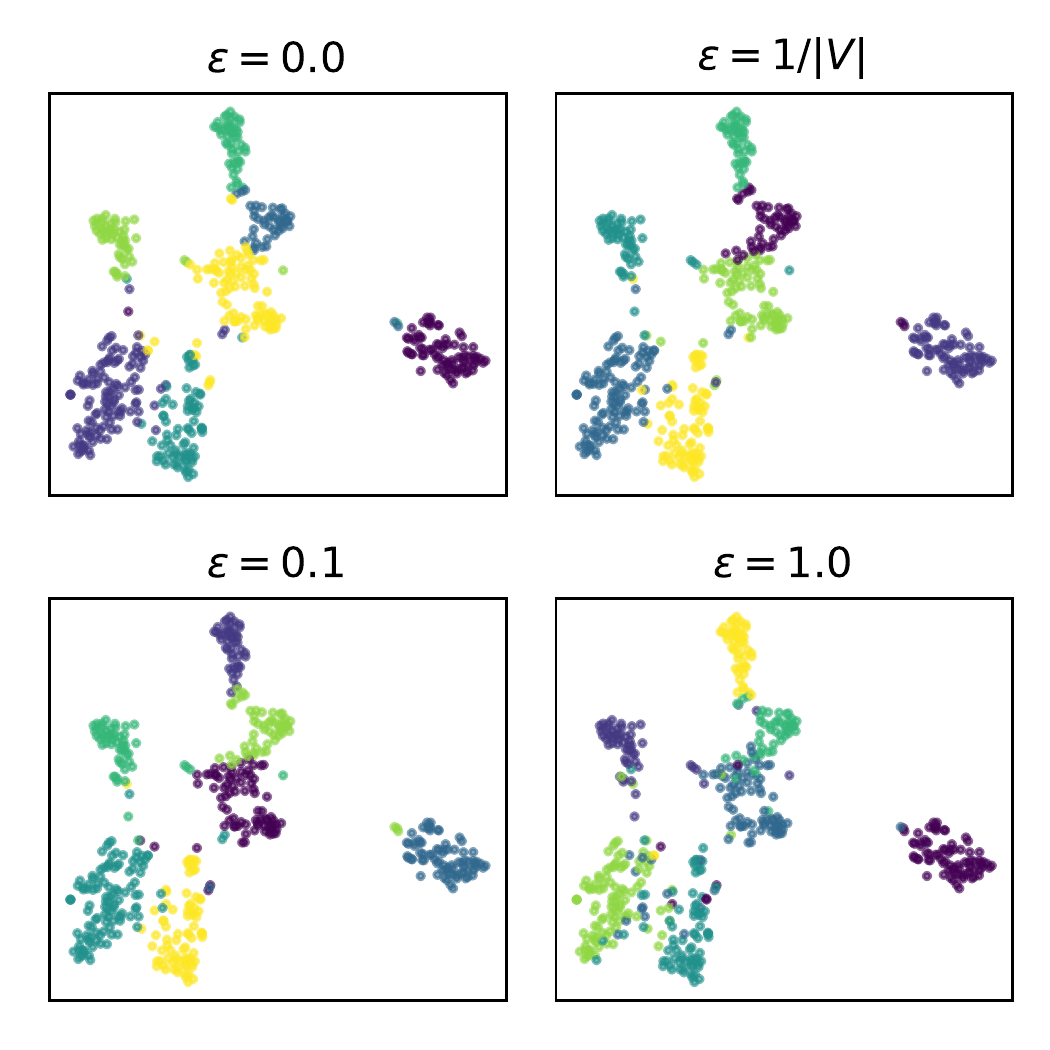}
   }
   \subfloat[2-hop clustering]{
     \includegraphics[width=0.4\textwidth]{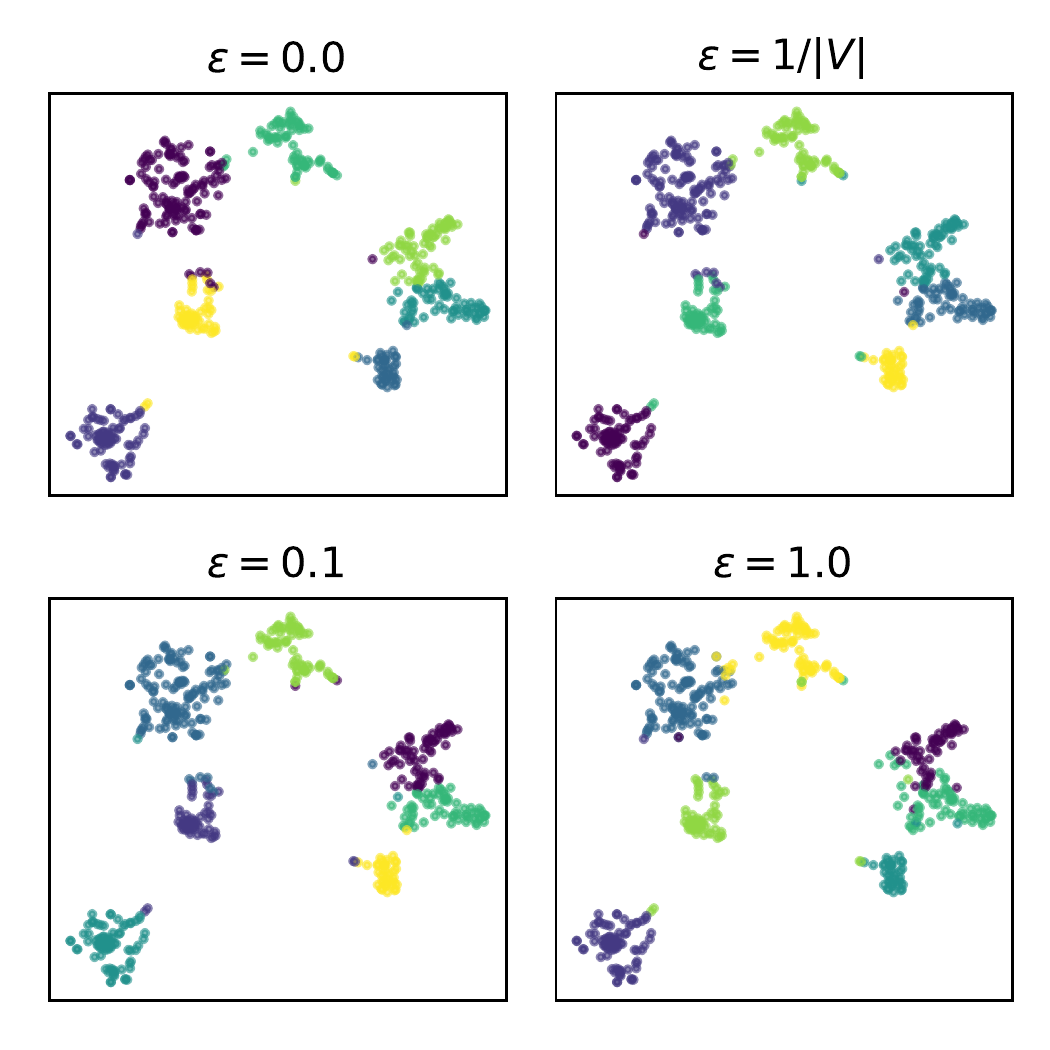}
   }
   \caption{Cora}
   \label{fig:cora_cluster_hop}
 \end{figure*}

 To compare the impact of graph sizes when applying the $\varepsilon$-LDP,
 Figure~\ref{fig:citeseer_cluster_hop} show the result from 1-hop and 2-hop subgraphs.
 where the pseudo label comes from the central node of the subgraph.
 As we can see, the GW based kmeans works better than the vanilla labeling.
 With a slight increase of $\varepsilon$, there will be a good performance in clustering.
 However, with a significant large privacy strength, it will introduce more deviates from the centroid.
 We also provide the clustering result from another dataset Cora in Figure~\ref{fig:cora_cluster_hop}.

\begin{figure*}
   \centering
   \subfloat[1-hop with diff. $\varepsilon$]{
     \includegraphics[width=0.45\textwidth]{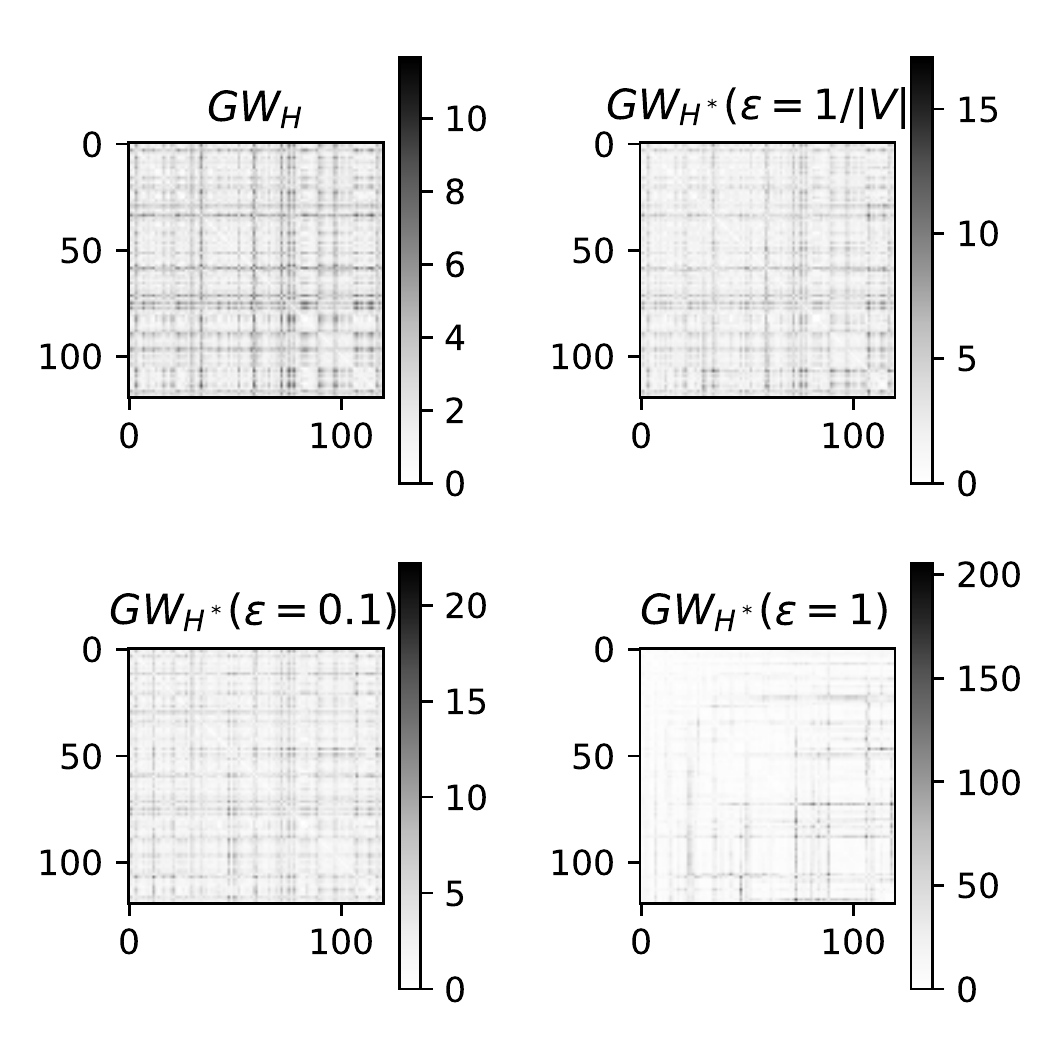}
   }
   \subfloat[2-hop with diff. $\varepsilon$]{
     \includegraphics[width=0.45\textwidth]{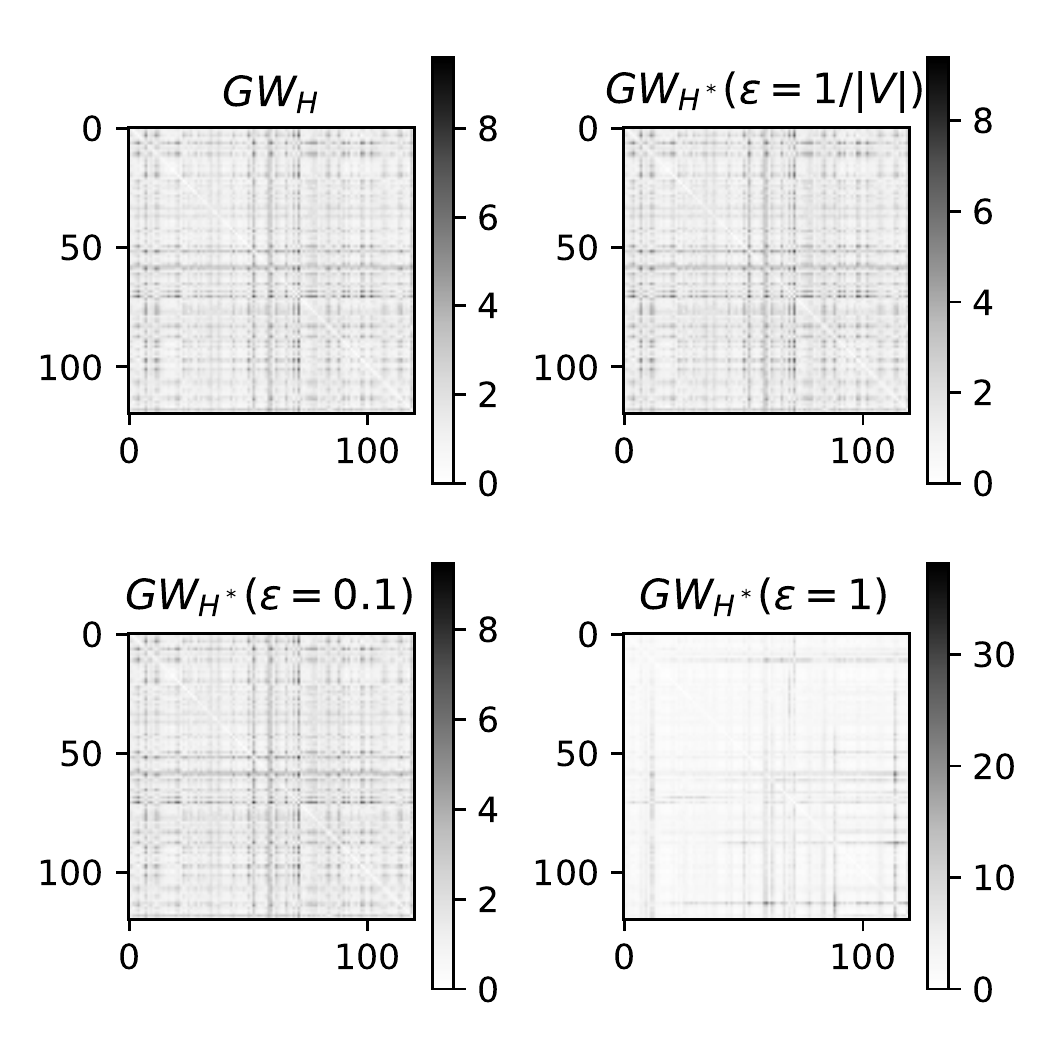}
   }
   \caption{Citeseer}
   \label{fig:ldp_citeseer_1_2}
 \end{figure*}

\begin{figure*}
   \centering
   \subfloat[1-hop with diff. $\varepsilon$]{
     \includegraphics[width=0.45\textwidth]{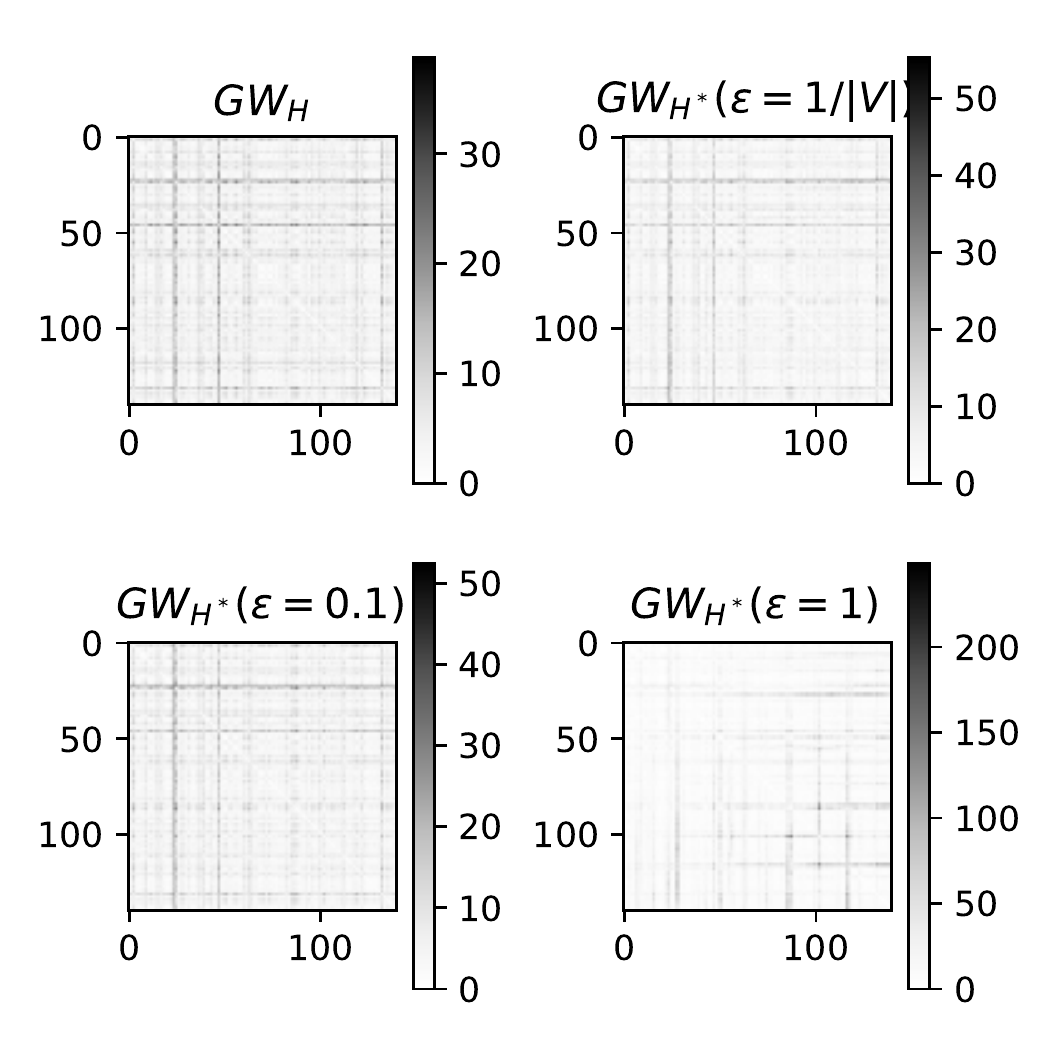}
   }
   \subfloat[2-hop with diff. $\varepsilon$]{
     \includegraphics[width=0.45\textwidth]{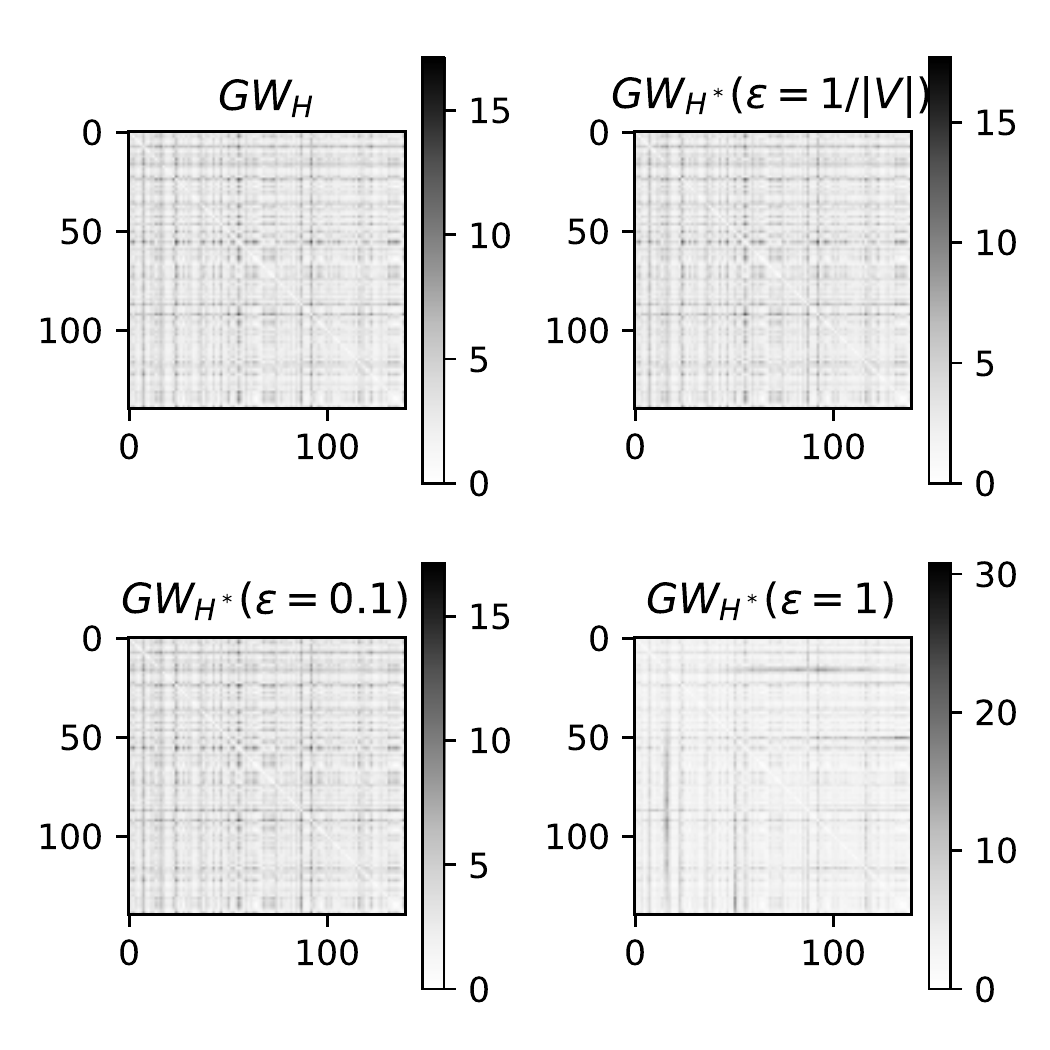}
   }
   \caption{Cora}
   \label{fig:ldp_cora_1_2}
 \end{figure*}

\begin{figure*}
   \centering
   \subfloat[1-hop with diff. $\varepsilon$]{
     \includegraphics[width=0.45\textwidth]{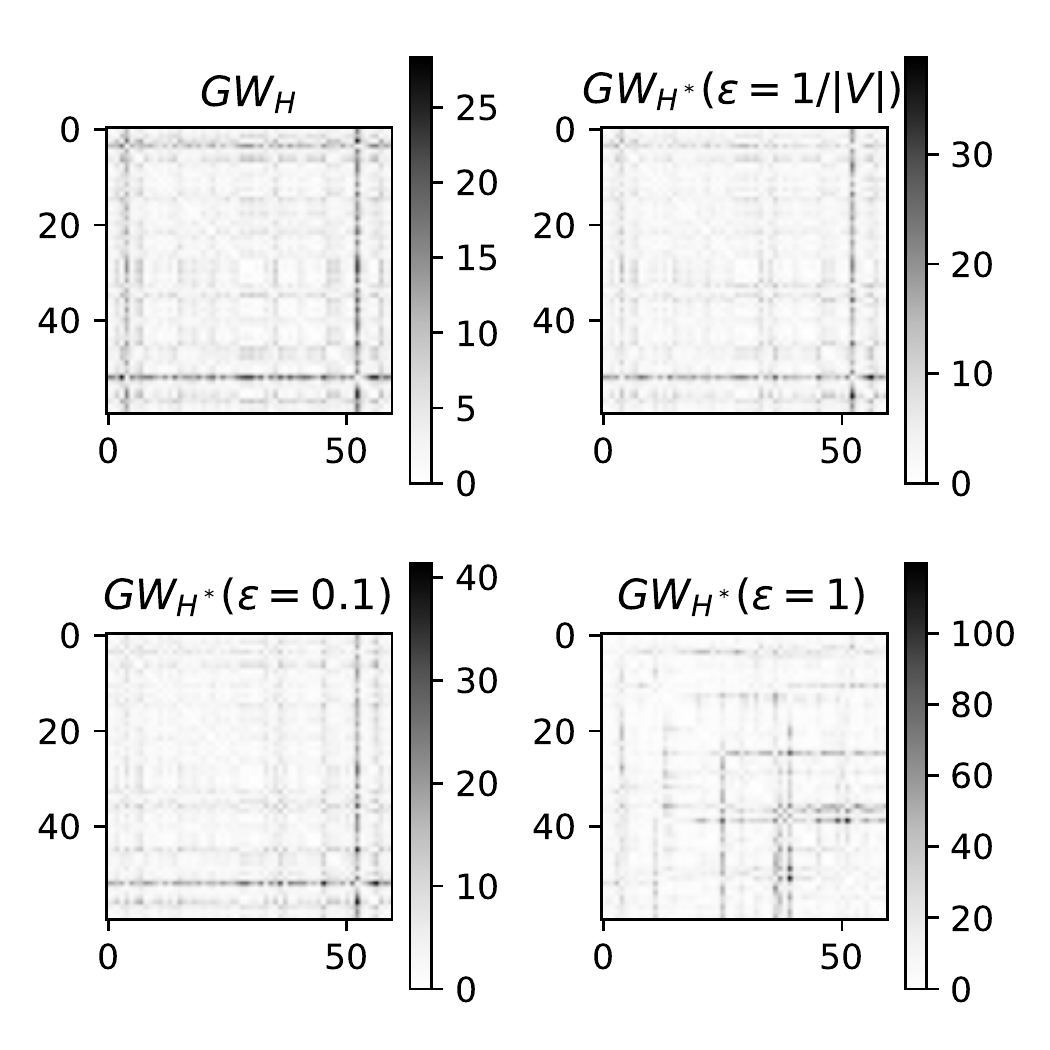}
   }
   \subfloat[2-hop with diff. $\varepsilon$]{
     \includegraphics[width=0.45\textwidth]{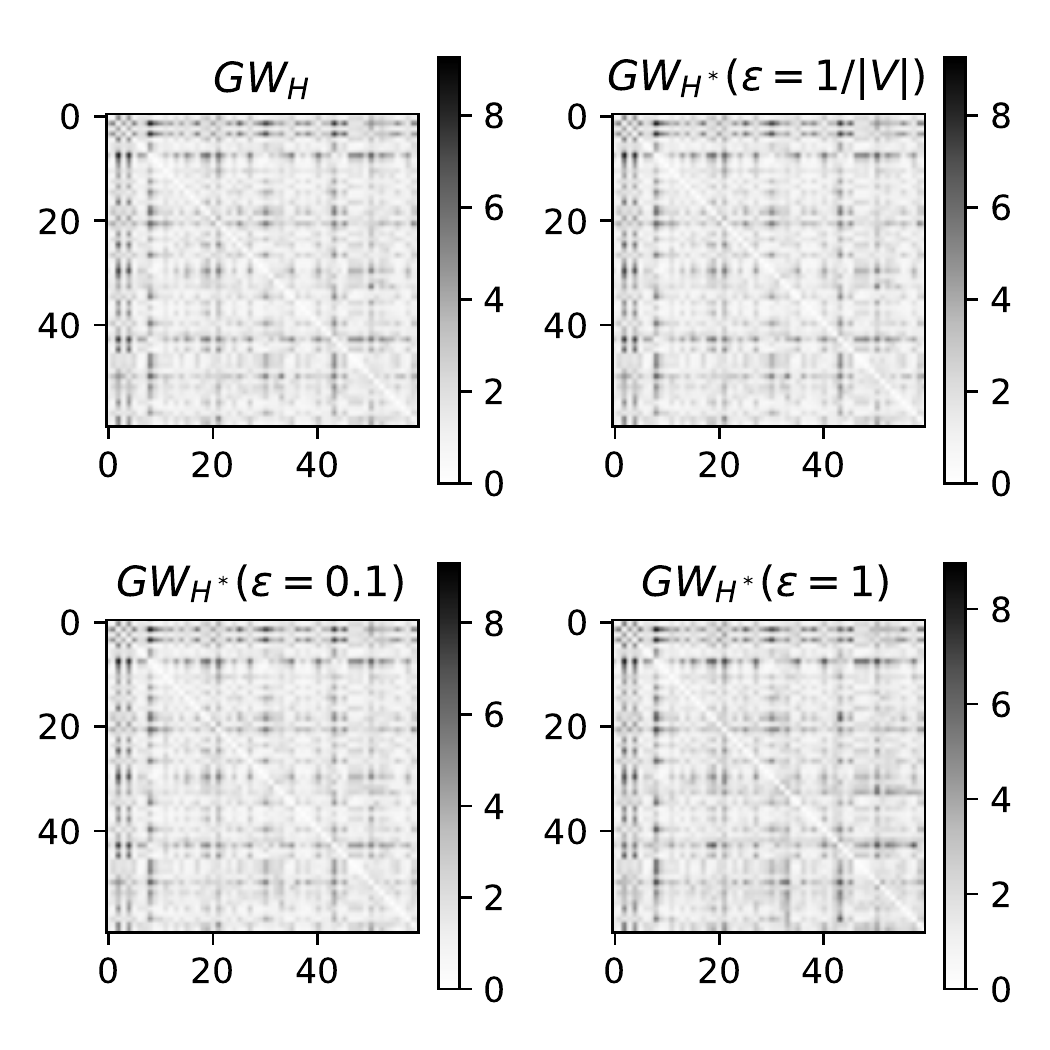}
   }
   \caption{Pubmed}
   \label{fig:ldp_pubmed_1_2}
 \end{figure*}

 For the GW metric with LDP in subgraphs,
 we take all the training nodes as the central node, and present GW metric with LDP under different buget with both 1-hop and 2-hop induced subgraphs. Figure~\ref{fig:ldp_citeseer_1_2},\ref{fig:ldp_cora_1_2},\ref{fig:ldp_pubmed_1_2} are the GW metric with LDP from citeseer, cora and pubmed, respectively.

 \begin{table}
   \centering
   \resizebox{0.5\textwidth}{!}{
     \begin{tabular}{|c|c|c|c|c|c|}
       \hline
       Dataset  & $\abs{\Vcal}$ & $\abs{\Ecal}$ & avg. degree & feature dim & num of labels \\
       \hline
       Citeseer & 2120          & 3679          & 3.47        & 3703        & 6             \\
       Cora     & 2485          & 5069          & 4.08        & 1433        & 7             \\
       Pubmed   & 19717         & 44324         & 4.50        & 500         & 3             \\
       \hline
     \end{tabular}
   }
   \caption{Single Graph Datasets}
   \label{tab:ds1}
 \end{table}

 \begin{table}
   \centering
   \resizebox{0.5\textwidth}{!}{
     \begin{tabular}{|c|c|c|c|c|c|}
       \hline
       Dataset     & num of graphs & node feature & num of label & avg nodes & avg edges \\
       \hline
       PROTEINS    & 1113          & 32           & 2            & 39.06     & 145.63    \\
       MUTAG       & 188           & 7            & 2            & 17.93     & 39.58     \\
       IMDB-Binary & 1000          & 0            & 2            & 19.77     & 193.06    \\
       \hline
     \end{tabular}
   }
   \caption{Multl-graph Dataset}
   \label{tab:ds2}
 \end{table}

 \begin{table}[h]
   \centering
   \resizebox{0.5\textwidth}{!}{
     \begin{tabular}{|c|c|c|c|c|}
       \hline
       Dataset                   & hop & avg. nodes & avg. edges & avg. degree \\
       \hline
       \multirow{2}{*}{Citeseer} & 1   & 4.61       & 5.15       & 1.8         \\
                                 & 2   & 25.13      & 44.26      & 2.58        \\
       \hline
       \multirow{2}{*}{Cora}     & 1   & 5.77       & 7.32       & 2.16        \\
                                 & 2   & 49.15      & 89.46      & 2.95        \\
       \hline
       \multirow{2}{*}{Pubmed}   & 1   & 5.95       & 6.65       & 1.66        \\
                                 & 2   & 53.83      & 103.95     & 2.82        \\
       \hline
     \end{tabular}
   }
   \caption{Statistics of split subgraphs}
   \label{tab:subgraph}
 \end{table}

\section{Similarity Measures.}
 \label{sec:similar_metric}
 To demonstrate the advantage of using Gromov-Wasserstein distance, we compare it with other graph metrics,
 \begin{figure*}[h]
   \centering
   \includegraphics[width=0.75\textwidth]{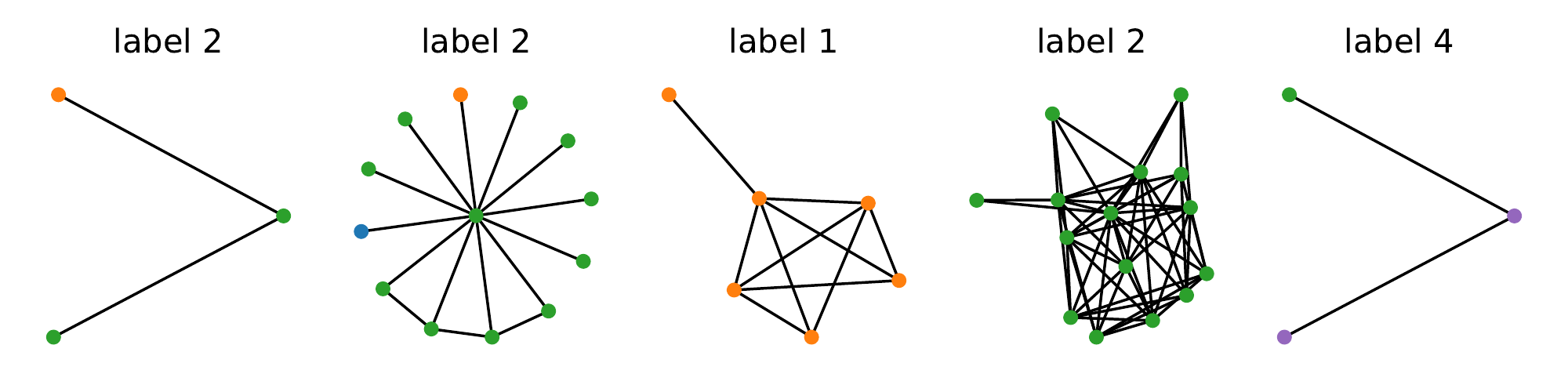}
   \includegraphics[width=0.75\textwidth]{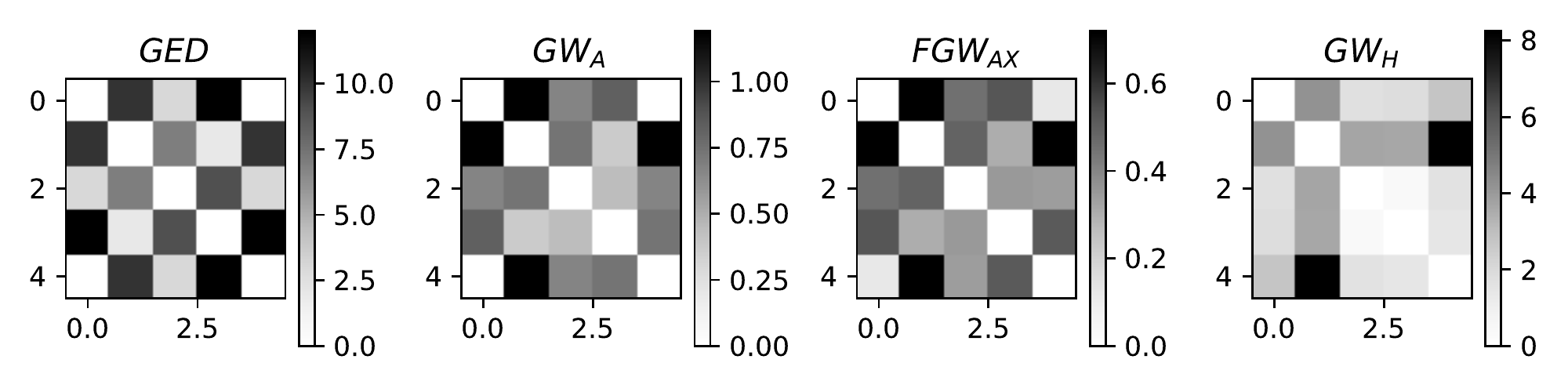}
   \caption{A demonstration of pairwise distance with different metrics.
     Five graphs in the upper row are the subgraph retrieved by 1-hop neighbors.
     Pairwise distances are presented in the second row.
     Both $GED$ and $GW_A$ are calculated from the structure information only, while $FGW_{AX}$ has both structure and feature involved when we set $\alpha=0.5$. $GW_H$ is the new pairwise distance evaluated from the node embedding.
   }
   \label{fig:metric_comp}
 \end{figure*}
 \begin{itemize}
   \item {\it Graph edit distance on topology}.
         Graph edit distance (GED) is a measure of similarity between two graphs in the fashion of their topology by summing over these optimal operations of insertion, deletion, and substitution for both nodes and edges.
         However, the exact calculation of GED is NP-hard \citep{zeng2009comparing}. Here, we adopt the exponential algorithm from \citet{abu2015exact} to calculate the GED for pairwise graphs.
   \item {\it Gromov-Wasserstein distance on topology}.
         We follow the setup in Eq.\ref{eq:gw_form2}, where optimal transportation is calculated based on graph topology. Regarding the structure distance $\Cvec$, they are computed by considering an all-pair shortest path (APSP) between the vertices.
   \item {\it Fused Gromov-Wasserstein distance on topology and node features}.
         Fused-GW ($\fgw$) \citep{titouan2019optimal} is an extension to the vanilla $\gw$, in order to cooperate with node feature as well.
         More specifically, it defines a trade-off parameter $\alpha \in \sbr{0, 1}$ between the feature and structure distance.
   \item {\it Gromov-Wasserstein discrepancy on node embedding}.
         Besides the distance calculate directly from the raw data, we also calculate the discrepancy based on the hidden representation as defined in Eq.\ref{eq:gw_h}.
 \end{itemize}
 Figure \ref{fig:metric_comp} provides an insightful view of the pairwise distance under different metrics.
 We randomly take five subgraphs induced from the 1-hop neighbors from Citeseer dataset, and the label from the central node is marked.
 Note that the first and second graphs have the same pseudo-label, and the first and the fifth graphs are isomorphic to each other.
 $GED$, $GW_A$, and $FGW_{AX}$ fail to detect similar properties from the first and second graphs.
 And for the isomorphism, thanks to the GW discrepancy, $GW_H$ is able to recognize the difference between those two graphs.
 Noting that $GED$, $GW_A$ has the topological information only, while  $FGW_{AX}$ and  $GW_H$ encode both structure and feature, while $GW_H$ has the additional learning process.

\section{Federate learning}
 \label{sec:fl}
 \paragraph{Server-side.} Under the federated learning setting,
 we take the aggregation algorithm $\mathsf{AGG}$ to build a global model in each round (line \ref{line:agg}) as shown in Algorithm \ref{alg:fedgnn_server}.
 Algorithms like FedAvg \citep{mcmahan2017communication}, FedProx \citep{li2020federated}, and MFL \cite{liu2020accelerating} can be applied accordingly.
 Moreover, it also allows the server to adopt secure aggregators \citep{bonawitz2016practical,so2020byzantine,li2019rsa} to improve the robustness of the global model.

 Besides the work of building the global model,
 we also explicitly enforce the server to construct a pairwise similarity matrix $\Dvec \in \RR^{N \times N}$ (line \ref{line:GW} in Algorithm \ref{alg:fedgnn_server}),
 which can be used for pipeline applications \citep{alvarez2018gromov,peyre2016gromov, demetci2020gromov}.
 $\Cvec_{\Hvec^{(i)}} \in \RR^{n^{(i)} \times n^{(i)}}$ represents intra-distance of the node embedding from $i$-th client,
 namely $\Hvec^{(i)} \in \RR^{n^{(i)} \times h}$, where $n^{(i)}$ is the number of node in $i$-th graph and $h$ is the hidden dimension.
 Therefore, we can explicitly formulate the distance matrix between node $u$ and $v$ in $\Cvec$ as
 \begin{align}
   \label{eq:gw_h}
   \Cvec_{\Hvec}(u, v) = d(\Hvec_u, \Hvec_v), \  \forall u, v \in \Vcal,
 \end{align}
 where $d(\cdot, \cdot)$ is the distance function applied.
 Due to its Euclidean space over the reals of $\Hvec$,
 we can simply take the Euclidean distance between each node embedding to build intra-distance.
 Of course, it can be easily extended to other distance metrics, such as cosine similarity and correlation.
 Noting that, after taking the distance introduced from hidden representation,
 the $\Cvec_{\Hvec}$ is a metric space induced from the original graph.
 Even though the unbalanced probability measure over nodes are also applicable \citep{titouan2019optimal},
 we will take $\pvec$ as a uniform distribution among the nodes, \ie, $\pvec_{u} = \frac{1}{\abs{\Vcal}}, \forall u \in \Vcal$.

 Additionally, the framework requires the optimization of Gromov-Wasserstein discrepancy on the server
 and can be retrieved only once after a stable global model.
 However, considering the unexpected drop-off of clients, we enforce the model to retrieve the encoded embedding periodically (parameter $s$).

 \paragraph{Client-side.} After get initial model from the server, the local client take full-batch training to update local model.
 $f(\cdot)$ is a shared GNN model with server.
 As we describe earlier, we also explicitly take the local differential privacy on node embeddings.
 By taking so, the raw data is protected, and also by taking the GNN, it is an implicit way to protect graph information.
 What's more, it secures the embedding with LDP privacy guaranteed,
 which again protect the raw data while sharing encoded embedding with the server.
 We provide our framework in Algorithm \ref{alg:fedgnn_server}, \ref{alg:fedgnn_local}.

 \begin{algorithm}[tb]
   \caption{Server's work}
   \label{alg:fedgnn_server}
   \textbf{Input}: Initial model $\wvec$ \\
   \textbf{Parameter}: periodic signal $s$, rounds $R$, number of clients $N$ \\
   \textbf{Output}: global model $\widetilde{\wvec}$, pairwise distance $\Dvec$
   \begin{algorithmic}[1] 
     \STATE $\widetilde{\wvec}_0 = \wvec$
     \FOR {$r$ in $1 \cdots R$}
     \FOR {$i$ in $1 \cdots N$}
     \STATE Send $\widetilde{\wvec}_{r-1}$ to clients
     \IF {$s$}
     \STATE $\wvec^{(i)}, \Hvec^{(i)} \leftarrow $ update on local client $i$
     \ELSE
     \STATE $\wvec^{(i)} \leftarrow $ update on local client $i$
     \ENDIF
     \ENDFOR
     \STATE $\widetilde{\wvec}_r \leftarrow \mathsf{AGG}(\wvec^{(i)}, \forall i \in N)$
     \label{line:agg}
     \STATE $\Dvec \leftarrow \mathsf{GW}(\Cvec_{\Hvec^{(i)}}, \Cvec_{\Hvec^{(j)}}, \pvec^{(i)}, \pvec^{(j)}), \forall i, j \in N^2$
     \label{line:GW}
     \ENDFOR
     \STATE \textbf{return} $\widetilde{\wvec}_r, \Dvec$
   \end{algorithmic}
 \end{algorithm}

 \begin{algorithm}[tb]
   \caption{Client's work}
   \label{alg:fedgnn_local}
   \textbf{Input}: Local graph $\Gcal (\Avec, \Xvec)$, global model $f(\widetilde{\wvec})$ \\
   \textbf{Parameter}: local epoch $E$, periodic signal $s$, LDP privacy $\varepsilon$\\
   \textbf{Output}: Updated model $\wvec^{(l)}$, node embedding $\Hvec^{(l)} \in \RR^{n \times h}$
   \begin{algorithmic}[1] 
     \STATE $\wvec_0 = \widetilde{\wvec}$
     \FOR{$e$ in $1 \cdots E$}
     \STATE Full-batch training to update $\wvec_e \leftarrow f(\Avec, \Xvec, \wvec_{e-1})$
     \ENDFOR
     \STATE $\wvec^{(l)} = \wvec_e$
     \IF {$s$}
     \STATE $\Hvec^{(l)} \leftarrow \mathsf{LDP}_G\rbr{f(\Avec, \Xvec, \wvec^{(l)}), \varepsilon}$
     \ENDIF
     \STATE \textbf{return} $\wvec^{(l)}, \Hvec^{(l)}$ if $s$.
   \end{algorithmic}
 \end{algorithm}

\section{Differential privacy on graphs}

 Differential privacy on graphs imposes a restriction of algorithms to protect the information of structured data.
 More specifically, such kind of algorithm helps to protect the statistics of graph, such as subgraph counts, number of edges and degree distribution \citep{kasiviswanathan2011can}.

 There are two kinds of differential privacy on graph, edge and node differential privacy.
 For the edge differential privacy, two graphs are neighbors if they differ exactly one edge, and the purpose is to protect the relation of entities in the graph.
 For the node differential privacy, two graphs are neighbors if they differ exact one node and its associated edges.
 We empirically study the difference when having an exact one edge/node difference.
 As mentioned earlier, LDP on topology is not a trivial roadmap in the GNN setting, especially the node DP has more influence on the difference due to the potential changes not only in nodes but also in its associated edges.



 \begin{figure}[h]
   \centering
   \includegraphics[width=0.35\textwidth]{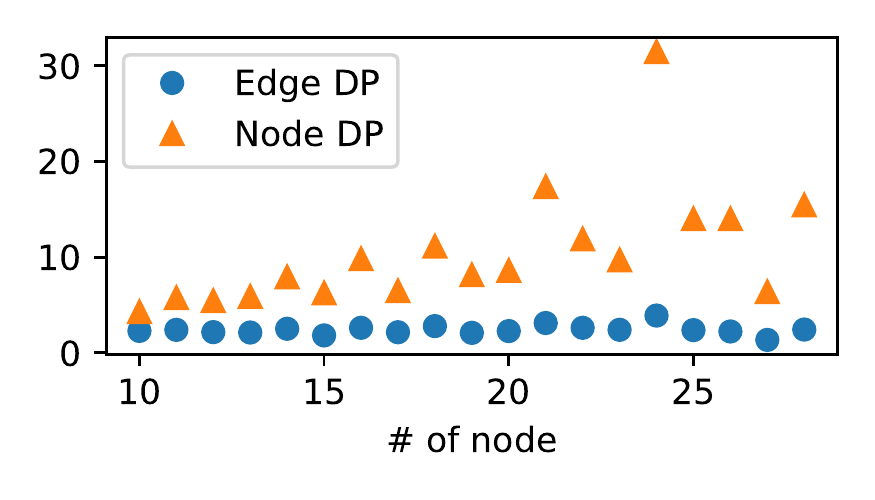}
   \caption{Difference in GW after exact one edge/node difference.}
   \label{fig:edge_node_dp_comp}
 \end{figure}
 More specifically, given a graph $\Gcal$, we are interested in how the GW changes when comparing two neighbor graphs,
 $d = \abs{\gw(\Gcal', \Gcal)}$,
 where $\Gcal'$ and $\Gcal$ differ only one edge or node.
 We take the MUTAG dataset TUDataset \cite{morris2020tudataset} as an example.
 For each sample graph in the dataset, we show that the directly topology differential privacy has a diverse difference, especially for the graph neighbors in terms of nodes.



\end{document}